\documentclass{article}

\newcommand{\state}{\mathbf{s}_t}
\newcommand{\statenext}{\mathbf{s}_{t+1}}

\newcommand{\action}{\mathbf{a}_t}
\newcommand{\actionsequence}{\mathbf{a}_{t-\tau:t-1}}
\newcommand{\actionsequencewhole}{\mathbf{a}_{t-\tau:t}}
\newcommand{\actionnext}{\mathbf{a}_{t+1}}
\newcommand{\policy}{\pi_{\theta}}
\newcommand{\seqmodel}{\phi_{\theta}}

\newcommand{\augmentedreward}{\tilde{r}(\state, \actionsequencewhole)}

\PassOptionsToPackage{numbers, compress}{natbib}

\usepackage[preprint]{neurips_2023}

\usepackage{natbib}
\usepackage{algorithm}
\usepackage{algpseudocode}
\usepackage{xcolor}
\usepackage{amsmath}
\usepackage{pythonhighlight}
\usepackage[utf8]{inputenc} 
\usepackage[T1]{fontenc}    
\usepackage{url}            
\usepackage{booktabs}       
\usepackage{amsfonts}       
\usepackage{nicefrac}       
\usepackage{microtype}      
\usepackage{xcolor}         
\usepackage{courierten}
\usepackage{listings}
\usepackage{graphicx, wrapfig}
\usepackage[colorlinks = true, urlcolor  = teal, citecolor=teal]{hyperref}
\definecolor{codegreen}{rgb}{0,0.6,0}
\definecolor{codegray}{rgb}{0.5,0.5,0.5}
\definecolor{codepurple}{HTML}{8570e0}
\definecolor{darkred}{HTML}{bd0404}
\definecolor{backcolour}{rgb}{1,1,1}

\lstdefinestyle{mystyle}{
    backgroundcolor=\color{backcolour},   
    commentstyle=\color{teal},
    keywordstyle=\bfseries\color{codepurple},
    numberstyle=\tiny\color{codegray},
    stringstyle=\color{codepurple},
    basicstyle=\ttfamily\footnotesize,
    breakatwhitespace=false,         
    breaklines=true,                 
    captionpos=b,                    
    keepspaces=true,                 
    numbers=left,                    
    numbersep=5pt,                  
    showspaces=false,                
    showstringspaces=false,
    showtabs=false,                  
    tabsize=2,
    comment=[l]{\#},
    keywords=[1]{len, policy, numpy, ravel, compress, concatenate, import, as, flatten, quantize, floor},
    keywords = [2]{def, return},
    keywordstyle=[2]\bfseries\color{darkred}
}
\title{Reinforcement Learning with Simple Sequence Priors}

\author{Tankred Saanum$^{1\dagger}$\\
  \And
  Noémi Éltető$^1$\\
  \And
  Peter Dayan$^{1, 2}$\\
  \And
  Marcel Binz$^1$\\
  \And
  Eric Schulz$^1$\\
  \AND \normalfont $^1$Max Planck Institute for Biological Cybernetics, \ \ $^2$University of Tübingen \\ $^\dagger$\texttt{tankred.saanum@tuebingen.mpg.de}}

\begin{document}

\maketitle

\begin{abstract}

Everything else being equal, simpler models should be preferred over more complex ones. In reinforcement learning (RL), simplicity is typically quantified on an action-by-action basis -- but this timescale ignores temporal regularities, like repetitions, often present in sequential strategies. We therefore propose an RL algorithm that learns to solve tasks with sequences of actions that are compressible. We explore two possible sources of simple action sequences: Sequences that can be learned by autoregressive models, and sequences that are compressible with off-the-shelf data compression algorithms. Distilling these preferences into sequence priors, we derive a novel information-theoretic objective that incentivizes agents to learn policies that maximize rewards while conforming to these priors. We show that the resulting RL algorithm leads to faster learning, and attains higher returns than state-of-the-art model-free approaches in a series of continuous control tasks from the DeepMind Control Suite. These priors also produce a powerful information-regularized agent that is robust to noisy observations and can perform open-loop control.


\end{abstract}

\section{Introduction}

\begin{wrapfigure}{l}{0.5\textwidth}

\includegraphics[width=0.5\textwidth]{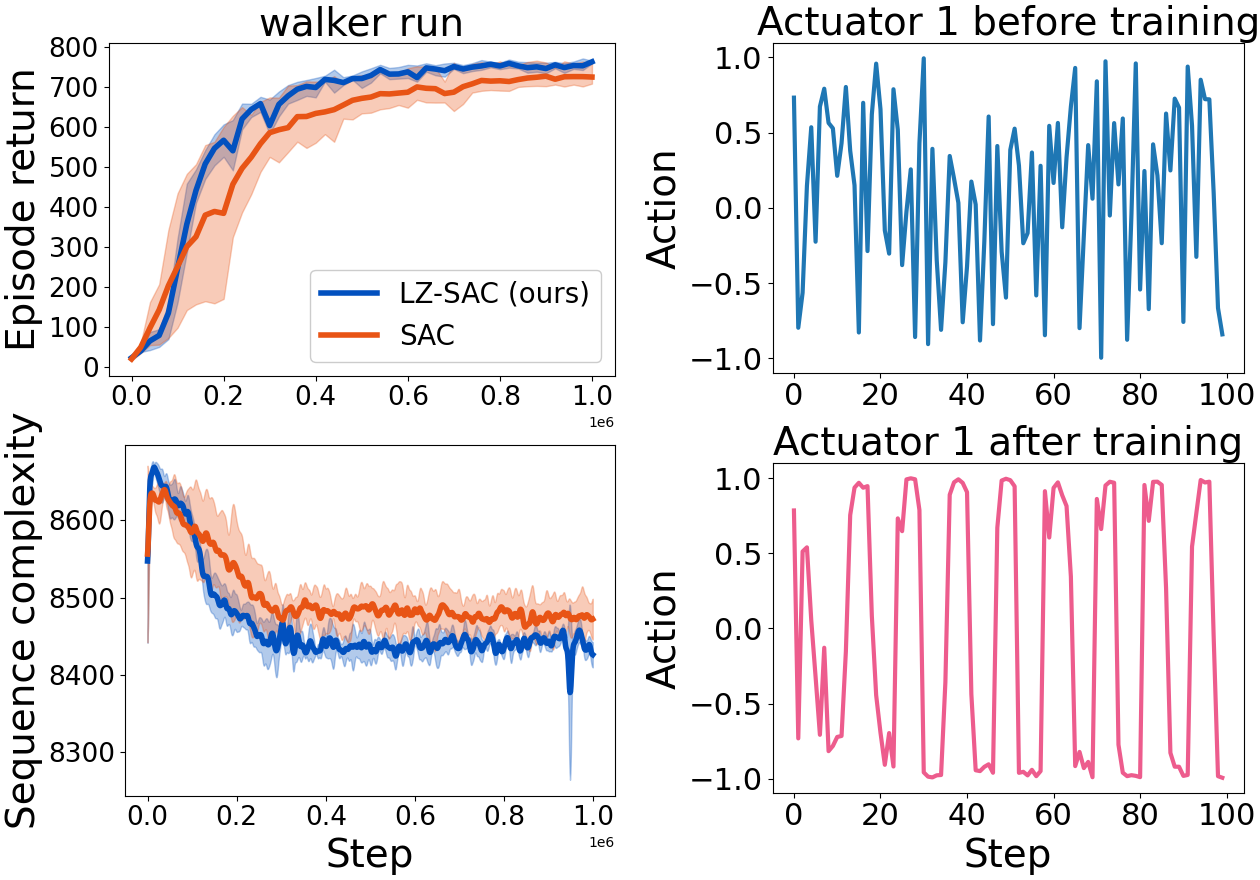}
\caption{Action sequences produced by a bipedal walker become  more compressible with learning. Our algorithm learns policies that solve tasks with simple action sequences, leading to decreased complexity and higher returns.}\label{fig:intro}
\end{wrapfigure}

Simplicity is a powerful inductive bias \cite{rasmussen2000occam, chater2003simplicity, solomonoff1964formal}. In science, we strive to build parsimonious theories and algorithms that involve repetitions of the same basic steps. Simplicity is also important in the context of reinforcement learning (RL). Policies that are simple are often easier to execute, and practical to implement even with limited computational resources \cite{eysenbach2021robust, ortega2013thermodynamics}. Many control problems have solutions that are compressible: Motor behaviors like running and walking involve moving our legs in a periodic, alternating fashion (Fig. \ref{fig:intro}). Here it is the sequence of actions selected that is compressible. Sequences with repetitive, periodic elements are easier to predict and can be compressed more than sequences that lack such structure. In the current work, we augment RL agents with a prior that their action sequences should be simple: If solutions to control problems are generally compressible, one should consider only the set of \emph{simple} solutions to a problem rather than the set of \emph{all} solutions. In a series of experiments, we show that RL with simple sequence priors produces policies that perform better and more robustly than state-of-the-art approaches without such priors.

\begin{figure}[t!]
\begin{center}
\includegraphics[width=1\textwidth]{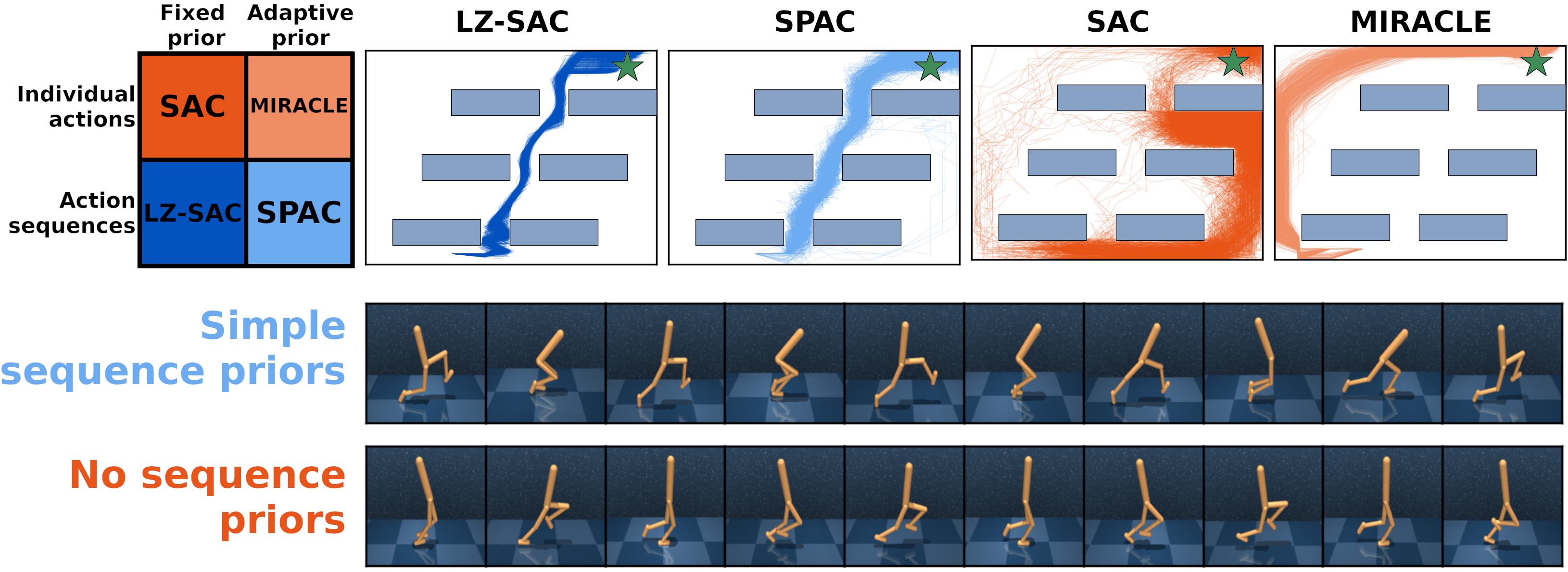}

\end{center}
\caption{\textbf{Top left:} Policy regularization either incentivizes sequences or individual actions to be close to the prior. Priors may be distinct in that they stay fixed over training or change from episode to episode with learning. \textbf{Top right:} Agents need to navigate to a goal location, where the shortest path requires fine control, following a repeating pattern. After learning, SAC randomly diffuses among multiple paths. MIRACLE prefers a simple path that only goes up and then to the right. Since the optimal path is compressible (repeating UP and RIGHT in a periodic fashion), the agents with the simple sequence priors prefer this path. \textbf{Bottom:} An agent with simple sequence priors, in this case SPAC, learns simple strategies for walking, using mostly the left leg to push itself forward in a repetitive fashion. }\label{fig:taxonomy}

\end{figure}

Though there are methods for regularizing policies with respect to the \emph{individual} actions they produce \cite{leibfried2020mutual, haarnoja2018soft}, we present a method that explicitly regularizes the \emph{sequences} of actions used to solve a task. Our regularization incentivizes the agent to use action sequences that can be compressed with a sequence prior. If an action sequence is likely under the prior, one needs fewer bits of information to represent it \cite{schmidhuber1992learning}. We explore two types of sequence priors: \textit{i}) Priors in the form of an autoregressive sequence model \cite{vaswani2017attention, hochreiter1997long} that learns to predict future actions based on actions that were performed in the past and \textit{ii}) priors distilled from a pre-programmed, lossless compression algorithm. Building on the Soft Actor-Critic algorithm (SAC) \cite{haarnoja2018soft}, we introduce Lempel-Ziv Soft Actor-Critic (LZ-SAC), using an off-the-shelf compression algorithm as its prior, and Soft Predictable Actor-Critic (SPAC), using a learned sequence prior (Fig. \ref{fig:taxonomy}).

The contributions of this paper are the following: We introduce a model-free RL algorithm for maximizing rewards with simple action sequences. In a series of continuous control tasks, we evaluate the utility of such simple sequence priors. First, we investigate whether simple sequence priors speed up policy search: In our experiments, agents with simple sequence priors consistently outperform state-of-the-art model-free RL algorithms in terms of reward maximization. This holds both in terms of learning speed and often in the final performance. Our second result is that our regularization produces an information-efficient RL agent, using fewer bits of information to solve control problems. Information-regularized models are more robust and better at generalizing \cite{alemi2016deep, eysenbach2021robust, zhang2020learning}. Lastly, we demonstrate the agents' advantages in environments with noisy and missing observations.\footnote{For videos showing behaviors learned with our algorithm, see our project website: \\ \url{https://sequencepriors.github.io}}

\section{Related work}

The idea of simplicity has received significant attention in previous work. Maximum entropy RL, for instance, augments the reward function with an entropy maximization term, effectively encouraging the agent to stay close to a simple uniform prior policy over actions \cite{ziebart2008maximum, levine2018reinforcement}. Many current approaches to deep RL -- such as SAC \cite{haarnoja2018soft} -- rely on this principle. This concept has been further extended by models like Mutual Information Regularized Actor-Critic Learning (MIRACLE) \cite{leibfried2020mutual} and others \cite{tishby2010information, grau2018soft}, which use a learnable state-independent prior policy instead of the uniform prior assumed by SAC. SAC and MIRACLE both induce simplicity at the level of individual actions. In contrast, our proposed approach works on the level of action sequences. 

It is not only possible to encode preferences for simplicity at the action level. Instead, simplicity can also be imposed by encouraging the agent to maintain simple internal representations -- the core idea behind the information bottleneck principle \cite{tishby2000information}. Deep RL agents that rely on this principle have many appealing properties, such as improved robustness to noise, better generalization, and more efficient exploration characteristics \cite{goyal2019infobot, igl2019generalization, lu2020dynamics}. Recently, \cite{eysenbach2021robust} demonstrated how to construct RL agents that learn policies that use few bits of information by not only compressing individual observations but entire sequences of observations. In some sense, our approach can be seen as a variant of the algorithm from \cite{eysenbach2021robust}. However, we compress sequences of \emph{actions}, rather than sequences of observations. Thus, our regularization does not target the complexity of the sequence of internal representations, but instead the complexity of the agent's behavior, manifested in the sequence of actions selected to solve a task.

Finally, simplicity is also an important feature of natural intelligence, where it has been repeatedly argued that simplicity is a unifying principle of human cognition \cite{chater2003simplicity}. For instance, \cite{lai2021policy} showed that people rely on compressed policies, ultimately leading to behavioral effects such as preservation or chunking \cite{wu2022learning, eltetHo2022tracking}. Likewise, \cite{binz2022modeling} demonstrated that human exploration behavior can be described by RL algorithms with limited description length, while \cite{kumar2022using} showed that compression captures human behavior in a visual search task.

\section{Control with simple sequences}\label{sec:methods}

In this section, we demonstrate how to construct RL agents that solve tasks using simple action sequences. We start by outlining the general problem formulation. We assume that the task can be posed as a Markov Decision Process (MDP). The MDP consists of a state space $\mathbf{s}\in\mathcal{S}$, an action space $\mathbf{a}\in \mathcal{A}$, and environment dynamics $p(\mathbf{s}_0)$ and $p(\statenext \mid \state, \action)$. The dynamics determine the probability of an episode starting in a particular state and the probability of the next state given the previous state and action, respectively. Lastly, there is the discount factor $\gamma$ and a reward function $r(\state, \action)$ that maps state-action pairs to a scalar reward term. The agent learns a policy $\policy(\action|\state)$ parameterized by $\theta$ that maps states to actions in a way that maximizes the sum of discounted rewards $\mathbb{E}_{\policy} \left[\sum_{t=1}^{T} \gamma^t r(\state, \action)\right]$. 

Though we want our RL agent to maximize rewards, we encourage it to do so with policies that produce simple action sequences. Inspired by previous approaches, we achieve this by augmenting the agent's objective \cite{haarnoja2018soft,eysenbach2021robust,levine2018reinforcement}, and search for a set of policy parameters $\theta$ that maximize reward while minimizing the \emph{complexity} of the policy $C(\actionsequencewhole, \state, \theta)$:

\begin{equation}
    \max_{\theta} \mathbb{E}_{\policy} \left[\sum_{t=1}^{T} \gamma^t (r(\state, \action) - \alpha\underbrace{C(\actionsequencewhole, \state, \theta)}_{\text{Complexity cost}}) \right ] 
\end{equation}

where the hyper-parameter $\alpha$ controls the trade-off between complexity and discounted rewards.

We can recover various previous approaches using this formulation. If we, for instance, set $C(\actionsequencewhole, \state, \theta) = \log \policy(\action\mid\state)$, we obtain maximum entropy RL algorithms such as SAC. SAC implicitly assumes a uniform prior over individual actions. An alternative to using the uniform prior in maximum entropy RL is to learn a parameterized prior over actions $p_{\theta}(\mathbf{a})$ based on the empirical distribution of actions the agent selects when solving the task \cite{grau2018soft, leibfried2020mutual}. Setting $C(\actionsequencewhole, \state, \theta) = \log \policy(\action\mid\state) - \log p_{\theta}(\action)$, we obtain MIRACLE.

\subsection{Simplicity with learned priors}
While both SAC and MIRACLE compress sums of \emph{individual} actions, they do not account for the structure that is present in whole action sequences. To close this gap, we present two methods for regularizing policies on the level of action sequences. For the first, we train a prior distribution $\seqmodel(\action\mid\actionsequence)$ to predict the agent's future actions from actions it performed in the past. We parameterize the prior as a neural sequence model. We use a causal transformer model \cite{vaswani2017attention, chen2021decision} to parameterize $\seqmodel$, though any type of sequence model could be used in principle. We can augment the reward function to incorporate the preference for predictable action sequences as follows:

\begin{equation}\label{eq:transformer}
    \tilde{r}(\state, \actionsequencewhole) = r(\state, \action) - \alpha(\log \policy(\action\mid \state) - \log \seqmodel(\action\mid\actionsequence))
\end{equation}

where $\actionsequence$ is a sequence of the last $\tau$ actions. Optimizing this objective, the agent will get rewarded for performing behaviors that the sequence model can predict better. The sequence model can learn to predict action sequences more easily if they contain structure and regularity. This has two interesting implications. \textit{i}) The agent is incentivized to visit states where its actions will be predictable, for instance by oscillating between states in a periodic manner. \textit{ii}) To perform actions that make it easier for the sequence model to predict future actions, for instance by performing behaviors that signal to the sequence model how it will behave in the future. We refer to this agent as the Soft Predictable Actor-Critic agent, or SPAC.

\subsection{Simplicity with compression algorithms}

\begin{wrapfigure}{r}{0.5\textwidth}

\includegraphics[width=0.5\textwidth]{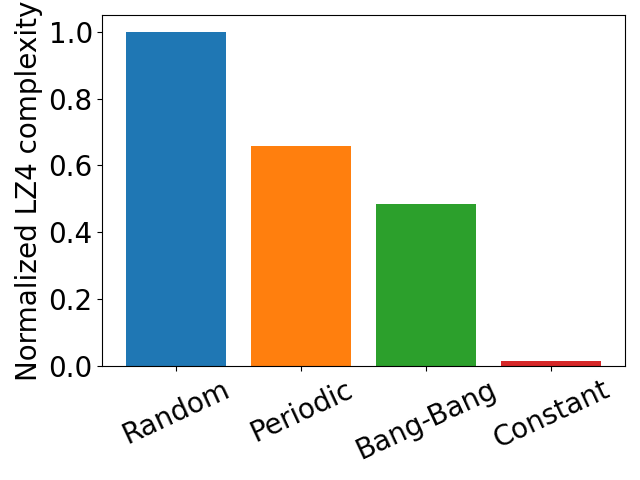}
\caption{Some sequences are more compressible than others. A sequence of randomly generated numbers is less compressible than sequences with periodicity, sequences that only contain two types of values (also known as Bang-Bang control), or constant sequences that only contain a single number.}\label{fig:lz4complexity}
\end{wrapfigure}

Since the sequence model and the policy are adapting their behavior and prior towards each other, the augmented reward function will change throughout training. This plasticity can make it challenging to search for viable policies. Moreover, training a sequence model on top of the RL agent creates additional computational overhead. We, therefore, explore the possibility of instilling a simplicity preference without the use of a sequence prior that necessarily adapts over episodes.

This second method for distilling simple sequence priors relies on off-the-shelf data compression algorithms \cite{sayood2017introduction}. Lossless data compression algorithms like \texttt{LZ4}, \texttt{bzip2} and \texttt{zlib} encode data into sequences of symbols from which the original data can be reconstructed or decompressed exactly. If there are repetitions, regularities, or periodicity in the data, the length of the encoded sequence can be significantly shorter than the original size of the data (Fig. \ref{fig:lz4complexity}). Relying on pre-programmed rules for data compression, this simplicity prior will not change over the course of training. Since compression algorithms like \texttt{LZ4} are fast, the sequence prior can be implemented with little computational overhead.

In this setting, we compute $C$ using the extra number of bits needed to encode $\action$ given that we have already encoded $\actionsequence$:

\begin{gather}
    \delta_t =  \text{len}(g(\actionsequence)) -\text{len}(g(\actionsequencewhole))\label{eq:delta}\\
    \tilde{r}(\state, \actionsequencewhole) = r(\state, \action) - \alpha (\log\policy(\action\mid\state) - \delta_t) \label{eq:zip}
\end{gather}
    
where $g(\cdot)$ is our compression function and $\text{len}(\cdot)$ returns the length of a sequence. We use the \texttt{LZ4} compression algorithm to compute the augmented rewards and refer to this agent as the LZ-SAC agent.

\subsection{Implementational details}

We implement all agents as extensions of the SAC algorithm. SAC is an off-policy actor-critic algorithm that performs maximum entropy RL. We train critics to learn the augmented $Q$-value function $\tilde{Q}(\state, \actionsequencewhole) = \mathbb{E}[\sum_{t=1}^N \gamma^t \augmentedreward]$ with temporal-difference learning \cite{sutton2018reinforcement}. The actors and sequence models are trained to minimize the same loss:
\begin{equation}
    \mathcal{L} =  \mathbb{E}_{\state, \actionsequencewhole \sim \mathcal{D}}[\alpha(\log \policy(\action\mid \state) - \log \seqmodel(\action\mid\actionsequence)) - \tilde{Q}(\state, \actionsequencewhole)]
\end{equation}

where $\mathcal{D}$ is a replay buffer and $\action\sim\policy(\cdot\mid\state)$. The LZ-SAC actor minimizes the same loss except that $- \log \seqmodel(\action\mid\actionsequence)$ is replaced with the term in Eq. \ref{eq:delta}. In practice, we take the minimum of two target $Q$-networks to train the actor and critic. Learning is achieved by sampling experiences from a replay buffer. To calculate the augmented rewards, we further sample action sequences $\actionsequence$ that led to the $(\state, \action, \statenext, r_t)$ tuple used for training (see Appendix \ref{A:implementation} for full implementational details).

\section{Simple sequence priors guide policy search}

\begin{figure}[h!]
\begin{center}
\includegraphics[width=1\textwidth]{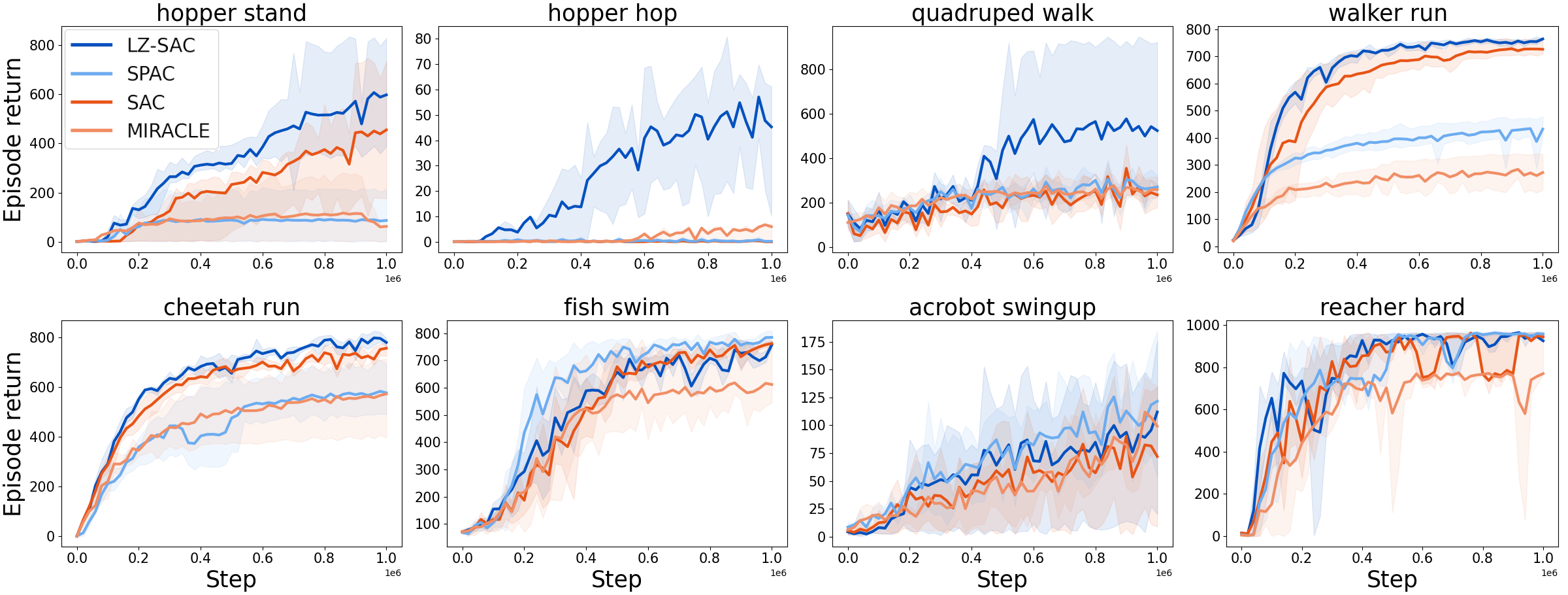}

\end{center}
\caption{Learning curves of agents in the DeepMind Control suite. Overall, LZ-SAC shows the best learning speed and final performance. Lines are the average episodic returns collected in 20 test episodes with a deterministic policy, averaged over five agents trained with different seeds. Shaded regions represent 20-80 performance percentiles.}\label{fig:main}

\end{figure}

We evaluated the four agents described in Section \ref{sec:methods} on eight continuous control tasks from the DeepMind Control Suite \cite{tassa2018deepmind}. We trained agents for 1 million environment steps across five seeds and evaluated their abilities at regular intervals with a deterministic policy, as in \citep{haarnoja2018soft, yarats2021improving}. We tuned $\alpha$ for each agent and found an $\alpha = 0.1$ to give the best performance in almost all tasks. Tasks and hyperparameter fitting is described in Appendix \ref{A:hyperparameters}.

In a majority of the tasks, the LZ-SAC agent outperforms the SAC and MIRACLE agents in learning speed and often final performance (Fig. \ref{fig:main}). At worst, the LZ-SAC agent matches the learning curves of SAC. This suggests that learning policies with simple sequence priors is indeed fruitful for policy search. We investigated whether this performance difference could simply be attributed to LZ-SAC acting more deterministically than SAC. Lowering the incentive of acting randomly for SAC did not close the performance gap, and often led to worse returns (see Appendix \ref{A:RS}).

In two tasks, \texttt{acrobot swingup} and \texttt{fish swim}, the SPAC agent shows a competitive advantage over the other models. However, the SPAC agent lags behind both the LZ-SAC agent and SAC agent in tasks from the \texttt{hopper}, \texttt{cheetah}, and \texttt{walker} domains. Here the policy that the SPAC agent learns achieves roughly 75\% of the return of the LZ-SAC agent.

\begin{wrapfigure}[18]{r}{0.5\textwidth}
\vspace{-10pt}
\includegraphics[width=0.5\textwidth]{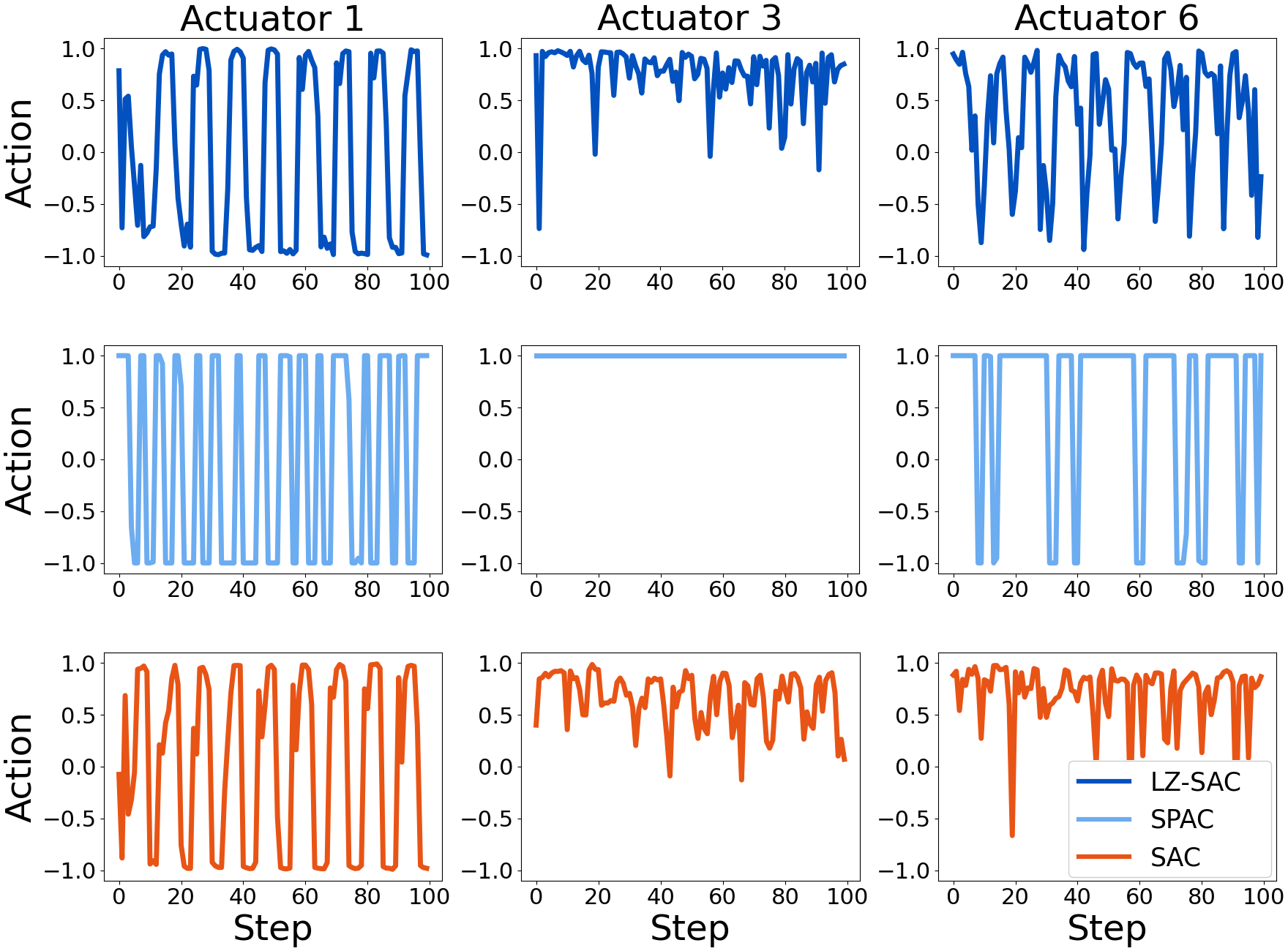}
\caption{Time series of actions produced by the LZ-SAC, SPAC and SAC agent in the \texttt{walker run} task. Action time series of the SPAC agent exhibits a simpler periodic pattern, even outputting a constant value for its third actuator. Actuators were chosen to show qualitatively different behaviors.}\label{fig:action_ts}
\end{wrapfigure}

The policies learned by the SPAC agent exhibit interesting properties: The agent has discovered solutions to these tasks that effectively use fewer action dimensions than the competitors (Fig. \ref{fig:action_ts}): For certain actuators $a_i$, the agent outputs a constant value throughout the episodes. For other actuators, the agent alternates between two extreme values, like a soft bang-bang controller \cite{bellman1956bang, seyde2021bang}. Essentially, the SPAC agent figures out which degrees of freedom it can eliminate without jettisoning rewards. Having fewer degrees of freedom makes it easier to predict the action sequences produced by the policy. This suggests that policy compression using adaptive sequence priors is better suited in tasks with low-dimensional action spaces.

\section{Simple sequence priors for information-regularized RL}



\begin{figure}[h!]
\begin{center}
\includegraphics[width=1\textwidth]{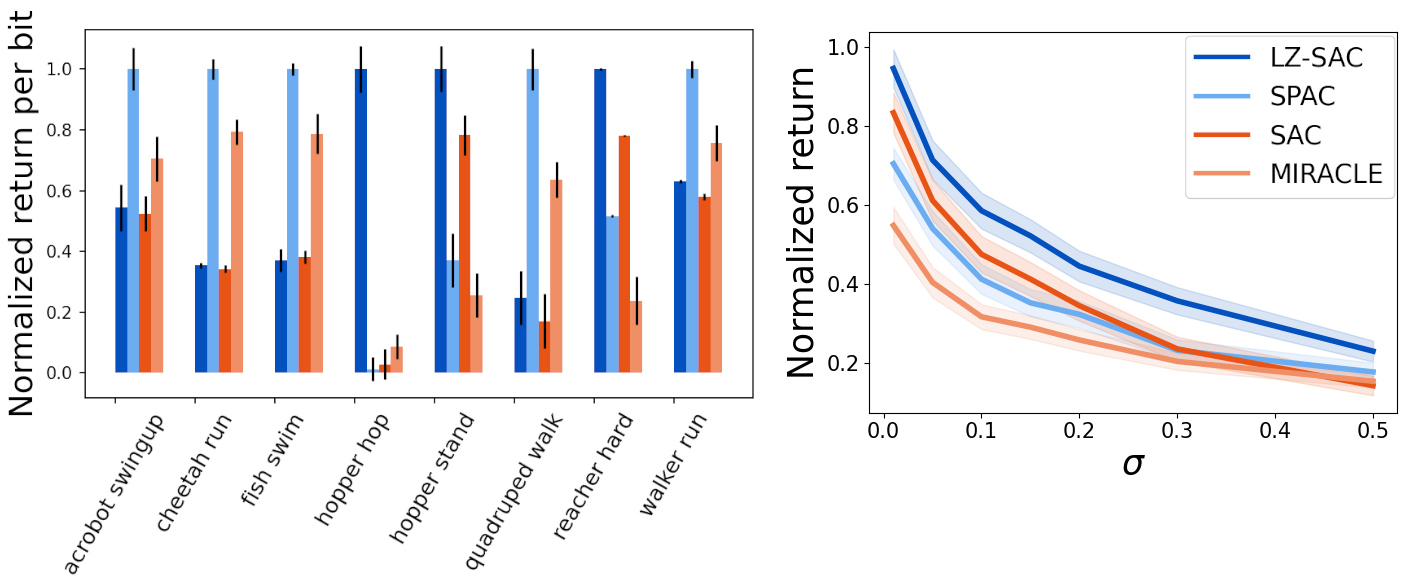}

\end{center}
\caption{\textbf{Left}: Normalized return per bit attained by the agents in the eight tasks. Agents with simple sequence priors achieve better return per bit ratios. Error bars represent the standard error of the mean (SEM). \textbf{Right}: Normalized return averaged over all tasks as a function of noise scale. Error bands represent the SEM.}\label{fig:RPB_noise}

\end{figure}

The expected difference in log-likelihood of the agents' actions under the policy versus the prior is an upper bound on the mutual information between states and actions \cite{eysenbach2021robust, alemi2016deep, tishby2000information}. Encouraging this difference to be low acts as an information-regularizer, the prior $p(\action|\actionsequence)$ being the information bottleneck. We tested the information-efficiency of learned policies; that is, how much reward the agents could collect relative to the information they used to make decisions. For the experiments, we again tested the deterministic versions of the agents. Simulating 50 episodes, we computed how much reward the agents were able to collect divided by the entropy of the distribution of actions used to solve the task $\mathbb{E}\left[\dfrac{\sum_{t=1}^T r_t}{\mathcal{H}[\mathbf{a}]}\right]$ (see Appendix \ref{A:RPB} for details and experiments with stochastic policies). Since the policies were deterministic, this entropy term approximated the mutual information between states and actions $I(\mathbf{s};\mathbf{a})$ (see Appendix \ref{A:MI}). In the left panel of Fig. \ref{fig:RPB_noise}, we show the normalized episodic return per bit. This quantity represents how much reward the agent attains per bit of information it uses on average to make a decision over the course of the episode.

The SPAC agent attains a superior return per bit ratio in five out of eight tasks. In the other three tasks, the LZ-SAC agent attains the highest return per bit ratio. This indicates that action sequence compression is a powerful information-regularizer, allowing agents to find policies that use significantly fewer bits of information to collect a certain amount of reward than the competitor approaches.

\section{Robustness to noise}



Information-regularized policies tend to show stronger robustness to noisy observations \cite{eysenbach2021maximum, eysenbach2021robust}: The less an agent's actions vary systematically with the state, the less will a perturbation to the agent's observation affect its actions. We assessed how observation noise affected the agents' ability to collect rewards. In the following experiments, we added Gaussian noise to the observations the agents used to execute their learned policies, $\state\leftarrow\state + \epsilon_t$ where $\epsilon_t\sim\mathcal{N}(\mathbf{0}, diag(\boldsymbol{\sigma}))$. We tested the agents on a series of noise scales $\sigma_j\in[0.01, 0.05, 0.1, 0.15, 0.2, 0.3, 0.5]$. The effect of noise was probed in all tasks except the \texttt{hopper hop} task, since here only the LZ-SAC agent reliably learned a policy that was better than random. Each agent was evaluated using 50 episodes for each noise-level.

We evaluated the agents based on how much reward they could collected given various levels of observation noise. Averaged over the tasks, the LZ-SAC agent showed the best ability to collect rewards when observations were perturbed with Gaussian noise (see the right panel in Fig. \ref{fig:RPB_noise}). The agents that were better at maximizing rewards showed a greater sensitivity to noise: LZ-SAC and SAC dropped to 28\% and 30\% of their average performance in the noise-free setting, respectively. While the LZ-SAC agent suffered greater percentage drops in return than the MIRACLE and SPAC agents, it still retained the highest performance for all noise levels. In the highest noise settings, SAC is comparable to the MIRACLE and SPAC agents, despite its generally stronger performance in the noise-free setting. This indicates that the LZ-SAC agent performed better in the noisy setting not only because the policy it learned was \emph{generally} better at maximizing rewards, but also because of robustness properties afforded by the sequence prior.

\section{Open-loop control}

If simple action sequences are pervasive in policies learned with RL, these priors could provide a good starting point for policy search. To further test this claim, we evaluated how well tasks from the DeepMind Control Suite could be solved by autoregressively generated action sequences from the sequence priors themselves. In our experiments, all agents produced the first 15 actions of an episode in a closed-loop manner. We then conditioned the sequence priors with these first 15 actions and sampled actions autoregressively for the remainder of the episode. The priors of the SAC and MIRACLE agents have no autoregressive component, and generated action sequences in a memory-less manner. We approximated samples from the \texttt{LZ4} prior by discretizing the action space and sampling the next action proportionally to how low its encoding cost is, given the previous actions.

\begin{figure}[h!]
\begin{center}
\includegraphics[width=1\textwidth]{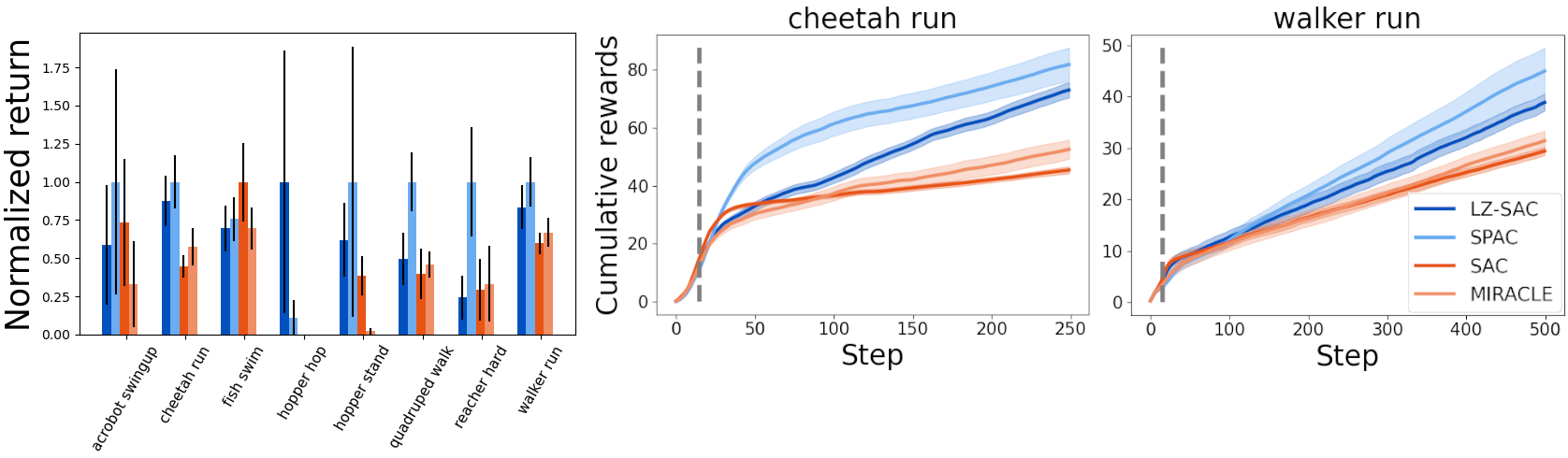}

\end{center}
\caption{\textbf{Left}: Bars represent return attained in the open-loop phase exclusively. Error bars represent the SEM. The sequence prior learned by the transformer generally performs the best. Notably, the \texttt{LZ4} prior performs well in tasks solved with periodic action sequences, like \texttt{cheetah} and \texttt{walker}. \textbf{Right}: Average cumulative reward obtained by agents in the \texttt{cheetah} and \texttt{walker} tasks. Dashed lines indicate where the open-loop controls start.}\label{fig:OL}

\end{figure}

The adaptive prior implemented as a transformer generally performs the best in the open-loop setting (Fig. \ref{fig:OL}, left). This is expected, as it was trained to predict behaviors that solve the tasks. In the \texttt{fish swim} task a uniform prior collects more rewards in the open-loop phase than the sequences generated by the transformer. However, increasing the number of closed-loop actions used to prompt the transformer to 25 made it surpass the performance of the uniform prior (Appendix \ref{A:OL}). This points to the importance of providing the sequence models with sufficient context to allow them to accurately predict behavior. 

More interesting is the performance of the prior obtained from the \texttt{LZ4} algorithm. Not only does it perform better than chance, but even comes close to the performance of the learned sequence prior in tasks like \texttt{cheetah run} and \texttt{walker run}. By conditioning on only a few actions from the policy, autoregressively approximating samples from \texttt{LZ4}'s prior produced behaviors outperforming the non-sequential priors used by SAC and MIRACLE (Fig. \ref{fig:OL}, right). This vindicates the compressibility prior as a starting point for policy search.

\section{Discussion}

We have argued that simplicity is a powerful principle to guide policy search in RL tasks. Because control problems are often solved with sequences of actions that contain repeating temporal patterns, we proposed to use simple sequence priors to create effective and robust RL agents. To provide agents with a notion of compressibility, we proposed two models: One where the strategy used for compression was fixed throughout training (LZ-SAC), and one where the strategy itself could change with experience (SPAC). While the LZ-SAC agents either outperformed or matched the performance of state-of-the-art methods like SAC, the SPAC agents learned more compressible strategies, attaining more rewards while using fewer bits of information to make a decision. Furthermore, agents trained with the LZ-SAC algorithm proved to be the most robust to observation noise. Lastly, both the trained transformer model and the prior distilled from the \texttt{LZ4} algorithm could autoregressively generate rewarding behaviors in continuous control tasks.

While SPAC showed a better ability to maximize rewards than MIRACLE, returns were lower than SAC and our alternative regularization technique. This is not unexpected. The transformer always required some amount of learning to be able to predict a particular action sequence. The \texttt{LZ4} algorithm, on the other hand, could immediately provide feedback about the compressibility of the agent's action sequences without any learning. For SPAC, having to learn a sequence prior induced a stronger bottleneck, resulting in more compressed policies. This is consistent with results reported by Eysenbach et al. \cite{eysenbach2021robust}, where a learned dynamics model was used to compress sequences of states: Here compression with a learned prior led to lower returns, but a higher return per bit rate. Our results suggest that sequence compression based on off-the-shelf compression algorithms like \texttt{LZ4} are better for policy search since there is no need for learning a sequence prior from scratch.



\textbf{Limitations:}
Action sequence compression requires either an adaptive prior, a neural sequence model, or a pre-programmed compression algorithm. The particular algorithm used for compression adds computational overhead and determines the types of action sequences that will be favored by the agent \cite{sayood2017introduction}. Future work should address the ways in which different compression algorithms or sequence priors affect policy regularization. Furthermore, a sufficiently sophisticated sequence model could in principle learn to predict complex action sequences. A possible extension of our work could be to further penalize the description length of the weights of the sequence model, or the compression algorithm, itself \cite{graves2011practical}. Finally, while we evaluated our algorithm on a large and diverse set of control tasks within the DeepMind Control Suite, the utility of simple sequence priors could be tested on other benchmarks. In discrete action settings, Atari games \cite{bellemare2013arcade} would be an appropriate benchmark.

\textbf{Future Directions:} A central feature of simple action sequences is that they are predictable. Being able to predict one's future behavior from past behavior could allow agents to simplify and compress their representations of the state of the world \cite{alemi2016deep, eysenbach2021robust}: If the point of observing the state is to determine what action to choose, one could discard information about the state of the world simply by considering the actions that were performed previously. This suggests that simple sequence priors could be beneficial for compressing policies and internal representations jointly.

\section*{Acknowledgements}
We thank the members of the Computational Principles of Intelligence Lab for feedback provided throughout the project, and Can Demircan for feedback on the manuscript. This work was supported by the Max Planck Society, the German Federal Ministry of Education and Research (BMBF): Tübingen AI Center, FKZ: 01IS18039A, and funded by the Deutsche Forschungsgemeinschaft (DFG, German Research Foundation) under Germany’s Excellence Strategy–EXC2064/1–390727645.
\newpage

\bibliography{refs}

\begin{thebibliography}{40}
\providecommand{\natexlab}[1]{#1}
\providecommand{\url}[1]{\texttt{#1}}
\expandafter\ifx\csname urlstyle\endcsname\relax
  \providecommand{\doi}[1]{doi: #1}\else
  \providecommand{\doi}{doi: \begingroup \urlstyle{rm}\Url}\fi

\bibitem[Rasmussen and Ghahramani(2000)]{rasmussen2000occam}
Carl Rasmussen and Zoubin Ghahramani.
\newblock Occam's razor.
\newblock \emph{Advances in neural information processing systems}, 13, 2000.

\bibitem[Chater and Vit{\'a}nyi(2003)]{chater2003simplicity}
Nick Chater and Paul Vit{\'a}nyi.
\newblock Simplicity: a unifying principle in cognitive science?
\newblock \emph{Trends in cognitive sciences}, 7\penalty0 (1):\penalty0 19--22,
  2003.

\bibitem[Solomonoff(1964)]{solomonoff1964formal}
Ray~J Solomonoff.
\newblock A formal theory of inductive inference. part i.
\newblock \emph{Information and control}, 7\penalty0 (1):\penalty0 1--22, 1964.

\bibitem[Eysenbach et~al.(2021)Eysenbach, Salakhutdinov, and
  Levine]{eysenbach2021robust}
Ben Eysenbach, Russ~R Salakhutdinov, and Sergey Levine.
\newblock Robust predictable control.
\newblock \emph{Advances in Neural Information Processing Systems},
  34:\penalty0 27813--27825, 2021.

\bibitem[Ortega and Braun(2013)]{ortega2013thermodynamics}
Pedro~A Ortega and Daniel~A Braun.
\newblock Thermodynamics as a theory of decision-making with
  information-processing costs.
\newblock \emph{Proceedings of the Royal Society A: Mathematical, Physical and
  Engineering Sciences}, 469\penalty0 (2153):\penalty0 20120683, 2013.

\bibitem[Leibfried and Grau-Moya(2020)]{leibfried2020mutual}
Felix Leibfried and Jordi Grau-Moya.
\newblock Mutual-information regularization in markov decision processes and
  actor-critic learning.
\newblock In \emph{Conference on Robot Learning}, pages 360--373. PMLR, 2020.

\bibitem[Haarnoja et~al.(2018{\natexlab{a}})Haarnoja, Zhou, Hartikainen,
  Tucker, Ha, Tan, Kumar, Zhu, Gupta, Abbeel, et~al.]{haarnoja2018soft}
Tuomas Haarnoja, Aurick Zhou, Kristian Hartikainen, George Tucker, Sehoon Ha,
  Jie Tan, Vikash Kumar, Henry Zhu, Abhishek Gupta, Pieter Abbeel, et~al.
\newblock Soft actor-critic algorithms and applications.
\newblock \emph{arXiv preprint arXiv:1812.05905}, 2018{\natexlab{a}}.

\bibitem[Schmidhuber(1992)]{schmidhuber1992learning}
J{\"u}rgen Schmidhuber.
\newblock Learning complex, extended sequences using the principle of history
  compression.
\newblock \emph{Neural Computation}, 4\penalty0 (2):\penalty0 234--242, 1992.

\bibitem[Vaswani et~al.(2017)Vaswani, Shazeer, Parmar, Uszkoreit, Jones, Gomez,
  Kaiser, and Polosukhin]{vaswani2017attention}
Ashish Vaswani, Noam Shazeer, Niki Parmar, Jakob Uszkoreit, Llion Jones,
  Aidan~N Gomez, {\L}ukasz Kaiser, and Illia Polosukhin.
\newblock Attention is all you need.
\newblock \emph{Advances in neural information processing systems}, 30, 2017.

\bibitem[Hochreiter and Schmidhuber(1997)]{hochreiter1997long}
Sepp Hochreiter and J{\"u}rgen Schmidhuber.
\newblock Long short-term memory.
\newblock \emph{Neural computation}, 9\penalty0 (8):\penalty0 1735--1780, 1997.

\bibitem[Alemi et~al.(2016)Alemi, Fischer, Dillon, and Murphy]{alemi2016deep}
Alexander~A Alemi, Ian Fischer, Joshua~V Dillon, and Kevin Murphy.
\newblock Deep variational information bottleneck.
\newblock \emph{arXiv preprint arXiv:1612.00410}, 2016.

\bibitem[Zhang et~al.(2020)Zhang, McAllister, Calandra, Gal, and
  Levine]{zhang2020learning}
Amy Zhang, Rowan McAllister, Roberto Calandra, Yarin Gal, and Sergey Levine.
\newblock Learning invariant representations for reinforcement learning without
  reconstruction.
\newblock \emph{arXiv preprint arXiv:2006.10742}, 2020.

\bibitem[Ziebart et~al.(2008)Ziebart, Maas, Bagnell, Dey,
  et~al.]{ziebart2008maximum}
Brian~D Ziebart, Andrew~L Maas, J~Andrew Bagnell, Anind~K Dey, et~al.
\newblock Maximum entropy inverse reinforcement learning.
\newblock In \emph{Aaai}, volume~8, pages 1433--1438. Chicago, IL, USA, 2008.

\bibitem[Levine(2018)]{levine2018reinforcement}
Sergey Levine.
\newblock Reinforcement learning and control as probabilistic inference:
  Tutorial and review.
\newblock \emph{arXiv preprint arXiv:1805.00909}, 2018.

\bibitem[Tishby and Polani(2010)]{tishby2010information}
Naftali Tishby and Daniel Polani.
\newblock Information theory of decisions and actions.
\newblock In \emph{Perception-action cycle: Models, architectures, and
  hardware}, pages 601--636. Springer, 2010.

\bibitem[Grau-Moya et~al.(2018)Grau-Moya, Leibfried, and Vrancx]{grau2018soft}
Jordi Grau-Moya, Felix Leibfried, and Peter Vrancx.
\newblock Soft q-learning with mutual-information regularization.
\newblock In \emph{International conference on learning representations}, 2018.

\bibitem[Tishby et~al.(2000)Tishby, Pereira, and Bialek]{tishby2000information}
Naftali Tishby, Fernando~C Pereira, and William Bialek.
\newblock The information bottleneck method.
\newblock \emph{arXiv preprint physics/0004057}, 2000.

\bibitem[Goyal et~al.(2019)Goyal, Islam, Strouse, Ahmed, Botvinick, Larochelle,
  Bengio, and Levine]{goyal2019infobot}
Anirudh Goyal, Riashat Islam, Daniel Strouse, Zafarali Ahmed, Matthew
  Botvinick, Hugo Larochelle, Yoshua Bengio, and Sergey Levine.
\newblock Infobot: Transfer and exploration via the information bottleneck.
\newblock \emph{arXiv preprint arXiv:1901.10902}, 2019.

\bibitem[Igl et~al.(2019)Igl, Ciosek, Li, Tschiatschek, Zhang, Devlin, and
  Hofmann]{igl2019generalization}
Maximilian Igl, Kamil Ciosek, Yingzhen Li, Sebastian Tschiatschek, Cheng Zhang,
  Sam Devlin, and Katja Hofmann.
\newblock Generalization in reinforcement learning with selective noise
  injection and information bottleneck.
\newblock \emph{Advances in neural information processing systems}, 32, 2019.

\bibitem[Lu et~al.(2020)Lu, Lee, Abbeel, and Tiomkin]{lu2020dynamics}
Xingyu Lu, Kimin Lee, Pieter Abbeel, and Stas Tiomkin.
\newblock Dynamics generalization via information bottleneck in deep
  reinforcement learning.
\newblock \emph{arXiv preprint arXiv:2008.00614}, 2020.

\bibitem[Lai and Gershman(2021)]{lai2021policy}
Lucy Lai and Samuel~J Gershman.
\newblock Policy compression: An information bottleneck in action selection.
\newblock In \emph{Psychology of Learning and Motivation}, volume~74, pages
  195--232. Elsevier, 2021.

\bibitem[Wu et~al.(2022)Wu, {\'E}lteto, Dasgupta, and Schulz]{wu2022learning}
Shuchen Wu, No{\'e}mi {\'E}lteto, Ishita Dasgupta, and Eric Schulz.
\newblock Learning structure from the ground up---hierarchical representation
  learning by chunking.
\newblock \emph{Advances in Neural Information Processing Systems},
  35:\penalty0 36706--36721, 2022.

\bibitem[{\'E}ltet{\H{o}} et~al.(2022){\'E}ltet{\H{o}}, Nemeth, Janacsek, and
  Dayan]{eltetHo2022tracking}
No{\'e}mi {\'E}ltet{\H{o}}, Dezs{\H{o}} Nemeth, Karolina Janacsek, and Peter
  Dayan.
\newblock Tracking human skill learning with a hierarchical bayesian sequence
  model.
\newblock \emph{PLoS Computational Biology}, 18\penalty0 (11):\penalty0
  e1009866, 2022.

\bibitem[Binz and Schulz(2022)]{binz2022modeling}
Marcel Binz and Eric Schulz.
\newblock Modeling human exploration through resource-rational reinforcement
  learning.
\newblock In \emph{Advances in Neural Information Processing Systems}, 2022.

\bibitem[Kumar et~al.(2022)Kumar, Correa, Dasgupta, Marjieh, Hu, Hawkins,
  Cohen, Narasimhan, Griffiths, et~al.]{kumar2022using}
Sreejan Kumar, Carlos~G Correa, Ishita Dasgupta, Raja Marjieh, Michael~Y Hu,
  Robert Hawkins, Jonathan~D Cohen, Karthik Narasimhan, Tom Griffiths, et~al.
\newblock Using natural language and program abstractions to instill human
  inductive biases in machines.
\newblock \emph{Advances in Neural Information Processing Systems},
  35:\penalty0 167--180, 2022.

\bibitem[Chen et~al.(2021)Chen, Lu, Rajeswaran, Lee, Grover, Laskin, Abbeel,
  Srinivas, and Mordatch]{chen2021decision}
Lili Chen, Kevin Lu, Aravind Rajeswaran, Kimin Lee, Aditya Grover, Misha
  Laskin, Pieter Abbeel, Aravind Srinivas, and Igor Mordatch.
\newblock Decision transformer: Reinforcement learning via sequence modeling.
\newblock \emph{Advances in neural information processing systems},
  34:\penalty0 15084--15097, 2021.

\bibitem[Sayood(2017)]{sayood2017introduction}
Khalid Sayood.
\newblock \emph{Introduction to data compression}.
\newblock Morgan Kaufmann, 2017.

\bibitem[Sutton and Barto(2018)]{sutton2018reinforcement}
Richard~S Sutton and Andrew~G Barto.
\newblock \emph{Reinforcement learning: An introduction}.
\newblock MIT press, 2018.

\bibitem[Tassa et~al.(2018)Tassa, Doron, Muldal, Erez, Li, Casas, Budden,
  Abdolmaleki, Merel, Lefrancq, et~al.]{tassa2018deepmind}
Yuval Tassa, Yotam Doron, Alistair Muldal, Tom Erez, Yazhe Li, Diego de~Las
  Casas, David Budden, Abbas Abdolmaleki, Josh Merel, Andrew Lefrancq, et~al.
\newblock Deepmind control suite.
\newblock \emph{arXiv preprint arXiv:1801.00690}, 2018.

\bibitem[Yarats et~al.(2021)Yarats, Zhang, Kostrikov, Amos, Pineau, and
  Fergus]{yarats2021improving}
Denis Yarats, Amy Zhang, Ilya Kostrikov, Brandon Amos, Joelle Pineau, and Rob
  Fergus.
\newblock Improving sample efficiency in model-free reinforcement learning from
  images.
\newblock In \emph{Proceedings of the AAAI Conference on Artificial
  Intelligence}, volume~35, pages 10674--10681, 2021.

\bibitem[Bellman et~al.(1956)Bellman, Glicksberg, and Gross]{bellman1956bang}
Richard Bellman, Irving Glicksberg, and Oliver Gross.
\newblock On the “bang-bang” control problem.
\newblock \emph{Quarterly of Applied Mathematics}, 14\penalty0 (1):\penalty0
  11--18, 1956.

\bibitem[Seyde et~al.(2021)Seyde, Gilitschenski, Schwarting, Stellato,
  Riedmiller, Wulfmeier, and Rus]{seyde2021bang}
Tim Seyde, Igor Gilitschenski, Wilko Schwarting, Bartolomeo Stellato, Martin
  Riedmiller, Markus Wulfmeier, and Daniela Rus.
\newblock Is bang-bang control all you need? solving continuous control with
  bernoulli policies.
\newblock \emph{Advances in Neural Information Processing Systems},
  34:\penalty0 27209--27221, 2021.

\bibitem[Eysenbach and Levine(2021)]{eysenbach2021maximum}
Benjamin Eysenbach and Sergey Levine.
\newblock Maximum entropy rl (provably) solves some robust rl problems.
\newblock \emph{arXiv preprint arXiv:2103.06257}, 2021.

\bibitem[Graves(2011)]{graves2011practical}
Alex Graves.
\newblock Practical variational inference for neural networks.
\newblock \emph{Advances in neural information processing systems}, 24, 2011.

\bibitem[Bellemare et~al.(2013)Bellemare, Naddaf, Veness, and
  Bowling]{bellemare2013arcade}
Marc~G Bellemare, Yavar Naddaf, Joel Veness, and Michael Bowling.
\newblock The arcade learning environment: An evaluation platform for general
  agents.
\newblock \emph{Journal of Artificial Intelligence Research}, 47:\penalty0
  253--279, 2013.

\bibitem[Haarnoja et~al.(2018{\natexlab{b}})Haarnoja, Zhou, Hartikainen,
  Tucker, Ha, Tan, Kumar, Zhu, Gupta, Abbeel,
  et~al.]{haarnoja2018softaplications}
Tuomas Haarnoja, Aurick Zhou, Kristian Hartikainen, George Tucker, Sehoon Ha,
  Jie Tan, Vikash Kumar, Henry Zhu, Abhishek Gupta, Pieter Abbeel, et~al.
\newblock Soft actor-critic algorithms and applications.
\newblock \emph{arXiv preprint arXiv:1812.05905}, 2018{\natexlab{b}}.

\bibitem[Hafner et~al.(2019)Hafner, Lillicrap, Ba, and
  Norouzi]{hafner2019dream}
Danijar Hafner, Timothy Lillicrap, Jimmy Ba, and Mohammad Norouzi.
\newblock Dream to control: Learning behaviors by latent imagination.
\newblock \emph{arXiv preprint arXiv:1912.01603}, 2019.

\bibitem[Nair and Hinton(2010)]{nair2010rectified}
Vinod Nair and Geoffrey~E Hinton.
\newblock Rectified linear units improve restricted boltzmann machines.
\newblock In \emph{Icml}, 2010.

\bibitem[Kingma and Ba(2014)]{kingma2014adam}
Diederik~P Kingma and Jimmy Ba.
\newblock Adam: A method for stochastic optimization.
\newblock \emph{arXiv preprint arXiv:1412.6980}, 2014.

\bibitem[Chung et~al.(2014)Chung, Gulcehre, Cho, and
  Bengio]{chung2014empirical}
Junyoung Chung, Caglar Gulcehre, KyungHyun Cho, and Yoshua Bengio.
\newblock Empirical evaluation of gated recurrent neural networks on sequence
  modeling.
\newblock \emph{arXiv preprint arXiv:1412.3555}, 2014.

\end{thebibliography}

\medskip

\newpage
\appendix
\section{Implementation details}\label{A:implementation}

Our algorithm is an extension of the Soft Actor-Critic algorithm \cite{haarnoja2018soft, haarnoja2018softaplications}, implemented in PyTorch. Like \cite{yarats2021improving, hafner2019dream}, we initialize agents' replay buffer with a 1000 seed observations collected with a uniform random policy. We update the Q-network pairs to predict the augmented $Q$ value function at every interaction step, and the actor network to maximize the augmented $Q$ value function every second interaction step. The augmented $Q$ targets take the following form:

\begin{equation}\label{eq:Qvals}
    y = r(\state, \action) + \gamma\left[ Q(\statenext, \actionnext) - \alpha(\log \policy(\actionnext|\statenext) - \log \seqmodel(\actionnext|\actionsequencewhole)) \right]
\end{equation}
where $\actionnext\sim\policy(\cdot|\statenext)$. To train the networks we sampled $B$ tuples of state, actions, next state, reward and terminal flags, as well as the $\tau$ actions that led to them. To allow the agent to train on observations early in episodes, we sampled $\tau$ from a uniform distribution of integers between 5 and $\tau_{\text{max}}$ (see tables \ref{table:hyperparameters}, \ref{table:transformer}) for every mini-batch sample used for training. We update the adaptive priors used in SPAC and MIRACLE together with the actor network. All actor and critic networks consisted of two hidden layers with 256 ReLU units \cite{nair2010rectified} each. The action prior used in MIRACLE was implemented as a multivariate isotropic Gaussian with learnable mean and standard deviation.

Since actions are bounded between -1 and 1, we transform actions sampled from the policy using the $\tanh$ transform $\action = \tanh(\mathbf{u}_t), \mathbf{u}_t\sim \policy$. We transformed the log-likelihood of an action under a Gaussian policy $\policy$ or action prior $\seqmodel$ using the following formula \cite{haarnoja2018soft, haarnoja2018softaplications}:



    


\begin{gather}
    \log \policy(\action|\state) = \log \mu(\mathbf{u}_t|\state) - \sum_{i=1}^D \log (1 - \tanh^2 (u_i))\\
    \log \seqmodel(\action|\actionsequence) = \log \psi_{\theta}(\mathbf{u}_t|\actionsequence) - \sum_{i=1}^D \log (1 - \tanh^2 (u_i))
\end{gather}

where $\log \psi_{\theta}(\mathbf{u}_t|\actionsequence)$ is the log likelihood of the untransformed action $\mathbf{u}_t$ under the untransformed sequence prior $\psi_{\theta}$.

\subsection{Quantifying compressibility}

We used the \texttt{LZ4} algorithm to quantify the compressibility of action sequences. The following code snippet describes how we computed the sequence complexity term in Eq. \ref{eq:delta}

\begin{lstlisting}[numbers=none]

sequence_i = action_sequence.numpy().ravel() 
# get length of compressed sequence at t
length_t1 = len(compression_algorithm.compress(sequence_i))
action_next = policy(state).numpy().ravel()  # get next on-policy action
sequence_j = np.concatenate((sequence_i, action_next), axis=0)
# get length of compressed sequence at t+1
length_t2 = len(compression_algorithm.compress(sequence_j))
delta = length_t - length_t2  # delta is the difference

\end{lstlisting}

Since the actions were continuous vectors, we quantized all action sequences with the following function:
\begin{lstlisting}[numbers=none]

def quantize(action_sequence, N=100):
    return (action_sequence*N).floor()
\end{lstlisting}

Here $N$ determines the granularity of the quantization, with lower $N$ producing more coarse-grained sequences. We set $N=100$ for our experiments. 

\subsection{Pseudo-code for \texttt{lZ4} algorithm}

The \texttt{lZ4} algorithm compresses sequences by replacing repeating sub-sequences in the data with references to an earlier occurring copy of the sub-sequence. These copies are maintained in a sliding window. Repeating sub-sequences are encoded as \textit{length-distance} pairs $(l, d)$, specifying that a set of $l$ symbols have a match $d$ symbols back in the uncompressed sequence. The following pseudo-code sketches compression implemented by \texttt{LZ4} \cite{sayood2017introduction}:
    

    

\begin{algorithm}[H]\label{algo:mfrl}
\caption{LZ4 pseudo-code}
\begin{algorithmic}
\Require Buffer size $b$, window size $w$, sequence $\mathbf{k}$
\State $t=0$
\State window $\leftarrow\langle \ \rangle$
\While{t $<$ len($\mathbf{k}$)}
    \State match $\leftarrow$ longest repeated occurrence in window found in $\mathbf{k}_{t:t+b}$
    \If{match exists}
        \State $d\leftarrow$ distance to start of match
        \State $l\leftarrow$ length of match
        \State $c\leftarrow$ symbol at $\mathbf{k}_{t+l}$
    \Else
        \State $d\leftarrow$ 0
        \State $l\leftarrow$ 0
        \State $c\leftarrow$ 0
    \EndIf
    \State output $(d, l, c)$
    \State start $\leftarrow \max(t-w+l, 0)$
    \State end $t\leftarrow t+l$
    \State window $\leftarrow \mathbf{k}_{\text{start}:\text{end}}$
    \State $t \leftarrow t + l + 1$
    
\EndWhile
\end{algorithmic}
\end{algorithm}

    

\section{Hyperparameters}\label{A:hyperparameters}

\subsection{Increasing reward scale}\label{A:RS}

\begin{figure}[h!]
\begin{center}
\includegraphics[width=1\textwidth]{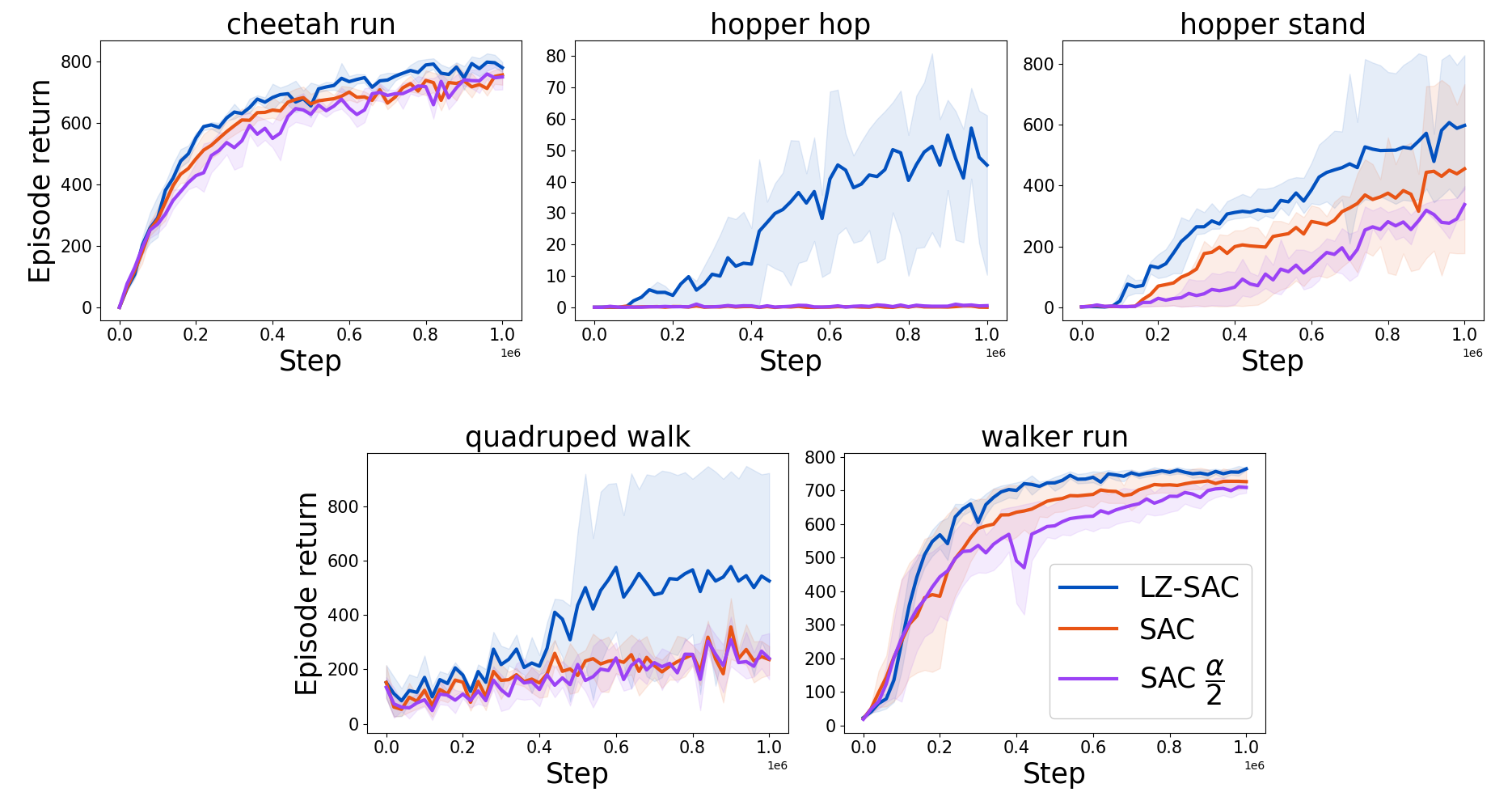}

\end{center}
\caption{Halving the incentive of acting randomly does not close the performance gap between LZ-SAC and SAC.}\label{fig:appendix}

\end{figure}

We investigated whether LZ-SAC's performance improvement could simply be attributed to the incentive to act more deterministically. We tested whether we could attain the same level of performance with SAC just by lowering the incentive of acting randomly. Doubling the scale of extrinsic reward relative to the intrinsic reward of acting randomly did not close the gap between the algorithms (Fig. \ref{fig:appendix}). Instead, we see a decline in performance when we set the incentive of randomness lower than $\alpha = 0.1$ (or $\alpha=0.02$ in \texttt{walker run}). This indicates that there is value in having a preference for simplicity on the sequence level that goes beyond simply being predictable at the level of individual actions.

\subsection{Algorithm hyperparameters}

We implement all algorithms using hyperparameters from \cite{yarats2021improving}, with slight deviations depending on the task. Since the computational overhead of using \texttt{lz4} as a compressor is small compared to the transformer, we train the agents using larger action sequences. All networks were trained with the Adam optimizer \cite{kingma2014adam}. A full list of hyperparameters is given below:

\begin{table}[h!]
\centering
\caption{Hyperparameters used for SAC, MIRACLE, LZ-SAC, and SPAC}\label{table:hyperparameters}
\begin{tabular}[t]{c|c}
\hline
\textbf{Hyperparameter} & \textbf{Value}\\
\hline
Complexity cost $\alpha$ & 0.1 \\
Complexity cost $\alpha$ (\texttt{walker run}) & 0.02 \\
Discount $\gamma$ & 0.99 \\
Critic update frequency & 1 \\
Actor update frequency & 2 \\
Action prior update frequency & 2 \\
Soft update $\rho$ & 0.01 \\
Batch size & 128 \\
Learning rate actor & $ 10^{-3}$ \\
Learning rate critic & $10^{-3}$ \\
Optimizer & Adam\\
Max context length LZ-SAC ($\tau_{\text{max}}$) & Interaction steps $\times$ 0.4 \\
Max context length LZ-SAC ($\tau_{\text{max}}$; \texttt{walker run}) & Interaction steps $\times$ 0.25 \\

\end{tabular}
\end{table}

\subsection{Transformer}

The SPAC agent uses a causal transformer \cite{vaswani2017attention} to learn a prior over action sequences. Our transformer was implemented with the following hyperparameters:

\begin{table}[h!]
\centering
\caption{Transformer hyperparameters.}\label{table:transformer}
\begin{tabular}[t]{c|c}
\hline
\textbf{Hyperparameter} & \textbf{Value}\\
\hline
Attention heads & 5 \\
Embedding dimensions & 30 \\
Learning rate decay & Linear \\
Warmup tokens & 10000 \\
Max context length ($\tau_{\text{max}}$) & 20 \\
Number of layers & 2 \\
Learning rate & $3 \times 10^{-4}$ \\
Dropout & 0.1 \\
Optimizer & Adam\\

\end{tabular}
\end{table}

\section{Task specification}\label{A:task}

We evaluated agents on tasks from the DeepMind Control Suite. Though dynamics are otherwise deterministic, the starting state of an episode is sampled from a distribution $p(\mathbf{s}_0)$. All episodes consist of 1000 environment steps. However, in practice the episode length is reduced to a number of \textit{interaction steps}, that is smaller than 1000. This is due to an action repeat hyperparameter which determines how many times an action $\action$ is repeated after it is selected. An action repeat value of 4 thus reduces the number of time steps where the agent needs to act to 250 interaction steps. The action repeat hyperparameter makes it more practical to train agents in the DeepMind Control Suite \cite{tassa2018deepmind}. We adopt conventional action repeat settings from the literature \cite{yarats2021improving}. In the \texttt{walker} and \texttt{hopper} domains we fitted the action repeat value for all agents among $[2, 4, 8]$ and chose the value that produced the best performance. Table \ref{action_repeat_values} shows the action repeat values used in our experiments:

\begin{table}[h!]
\centering
\caption{Action repeat values.}
\label{action_repeat_values}
\begin{tabular}[t]{c|c}
\hline
\textbf{Task} & \textbf{Action repeat}\\
\hline
\texttt{acrobot swingup} & 8 \\
\texttt{cheetah run} & 4 \\
\texttt{fish swim} & 4 \\
\texttt{hopper hop} & 8 \\
\texttt{hopper stand} & 8 \\
\texttt{quadruped walk} & 4 \\
\texttt{reacher hard} & 4 \\
\texttt{walker run} & 2 \\
\texttt{walker run} (SPAC) & 4 \\

\end{tabular}
\end{table}

\section{Mutual information approximation}\label{A:MI}

The mutual information $I(X;Y)$ between variables $X$ and $Y$ is a measure of how much they depend on each other. In our case we are interested in the mutual information between states and actions $I(\mathbf{s};\mathbf{a})$. The mutual information here quantifies how many bits of information knowing the outcome of the random variable $\state$ provides about the other random variable $\action$, in other words, how much the state reveals about what action will be selected. The more an agent's actions vary as a function of the state, the more bits of information the state reveals about the action that the agent will select.

The mutual information is defined as the following quantity

\begin{equation}
    I(\mathbf{a};\mathbf{s}) = \mathcal{H}[\mathbf{a}] - \mathcal{H}[\mathbf{a}|\mathbf{s}]
\end{equation}

\begin{wrapfigure}[18]{r}{0.5\textwidth}
\vspace{-15pt}
\includegraphics[width=.5\textwidth]{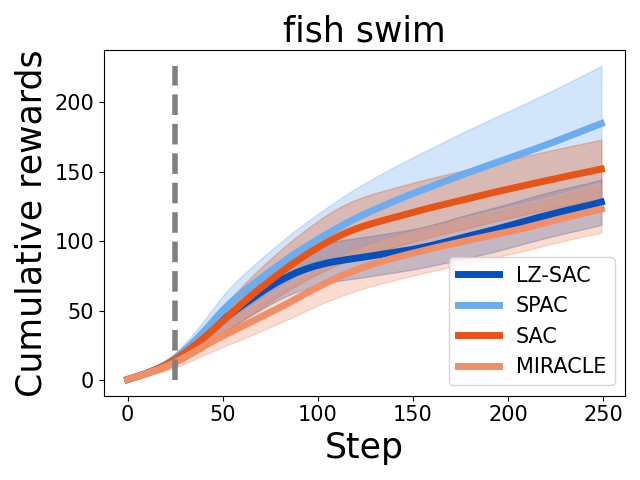}

\caption{Cumulative rewards attained by sequence priors in the \texttt{fish swim} environment, with a prompt length of 25.}\label{fig:appendix_OL}
\end{wrapfigure}

To compute the mutual information we need to know the entropy of the distribution of actions used to solve the task, and the conditional entropy of $\mathbf{a}|\mathbf{s}$. Since we make the policies deterministic, we know that $\mathcal{H}[\mathbf{a}|\mathbf{s}] = 0$. This way the mutual information reduces to the entropy over actions $\mathcal{H}[\mathbf{a}]$. Given a sample of actions produced by the agent solving the task, we approximate $\mathcal{H}[\mathbf{a}]$ the following way: We first quantized each selected action into $100\times|\mathcal{A}|$ bins, where $\mathcal{A}$ is the action space. We then calculated a categorical distribution over actions based on the frequencies of the quantized actions, the entropy of which we used as our approximation for $\mathcal{H}[\mathbf{a}]$. The categorical distribution was calculated based on actions selected over 50 episodes.

\subsection{Return per bit for stochastic policies}\label{A:RPB}

We evaluated the information efficiency of the stochastic variants of the policies learned by LZ-SAC, SPAC, SAC and MIRACLE. We approximated the entropy of the distribution of actions in the same way described above, sampling actions over 50 episodes. To compute the conditional entropy of actions given the state $\mathcal{H}[\mathbf{a}|\mathbf{s}]$, we sampled 1000 actions from the policy at every state $\state$. We then calculated a categorical distribution (described in previous section) based on this sample, the entropy of which we used as our approximation of the conditional entropy $\mathcal{H}[\mathbf{a}|\mathbf{s}]$. 

\begin{figure}[t!]
\begin{center}
\includegraphics[width=.65\textwidth]{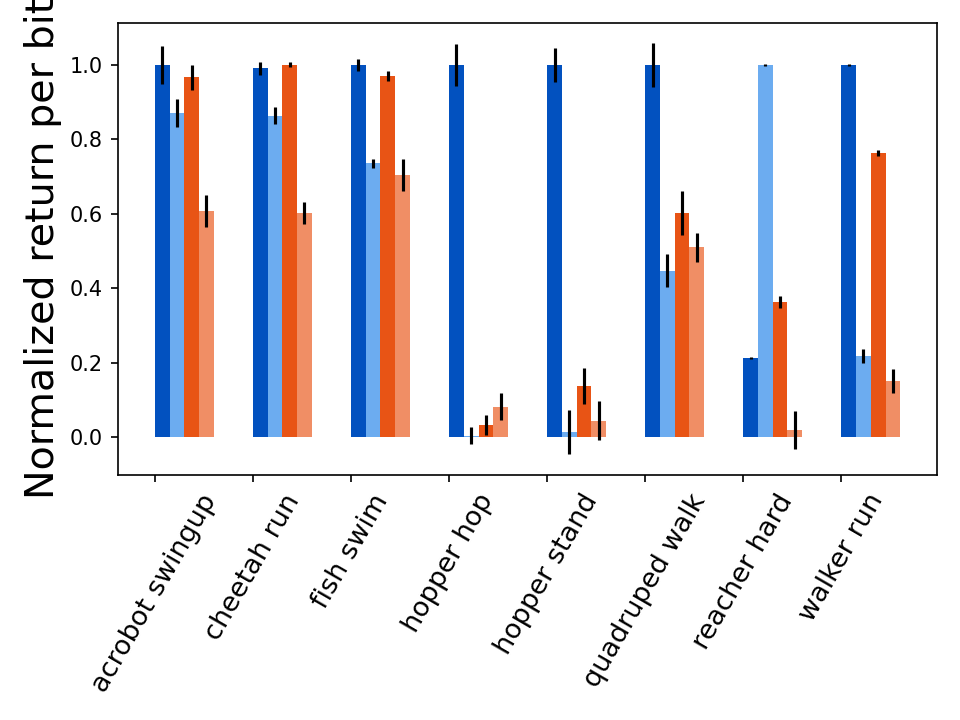}

\end{center}
\caption{Return per bit for the stochastic policies. LZ-SAC attains the highest return per bit ratio in most tasks.}\label{fig:appendix_rpb}

\end{figure}




In this setting, the LZ-SAC algorithm tends to produce the most information efficient agents (Fig. \ref{fig:appendix_rpb}). 

\section{Open-loop control}\label{A:OL}

Increasing the number of closed-loop actions used to prompt the transformer makes it generate more rewarding action sequences. This shows the importance of providing the sequence models with enough context, to be able to predict rewarding behaviors (Fig. \ref{fig:appendix_OL}).

\section{Partial observability}

\begin{figure}[h!]
\begin{center}
\includegraphics[width=\textwidth]{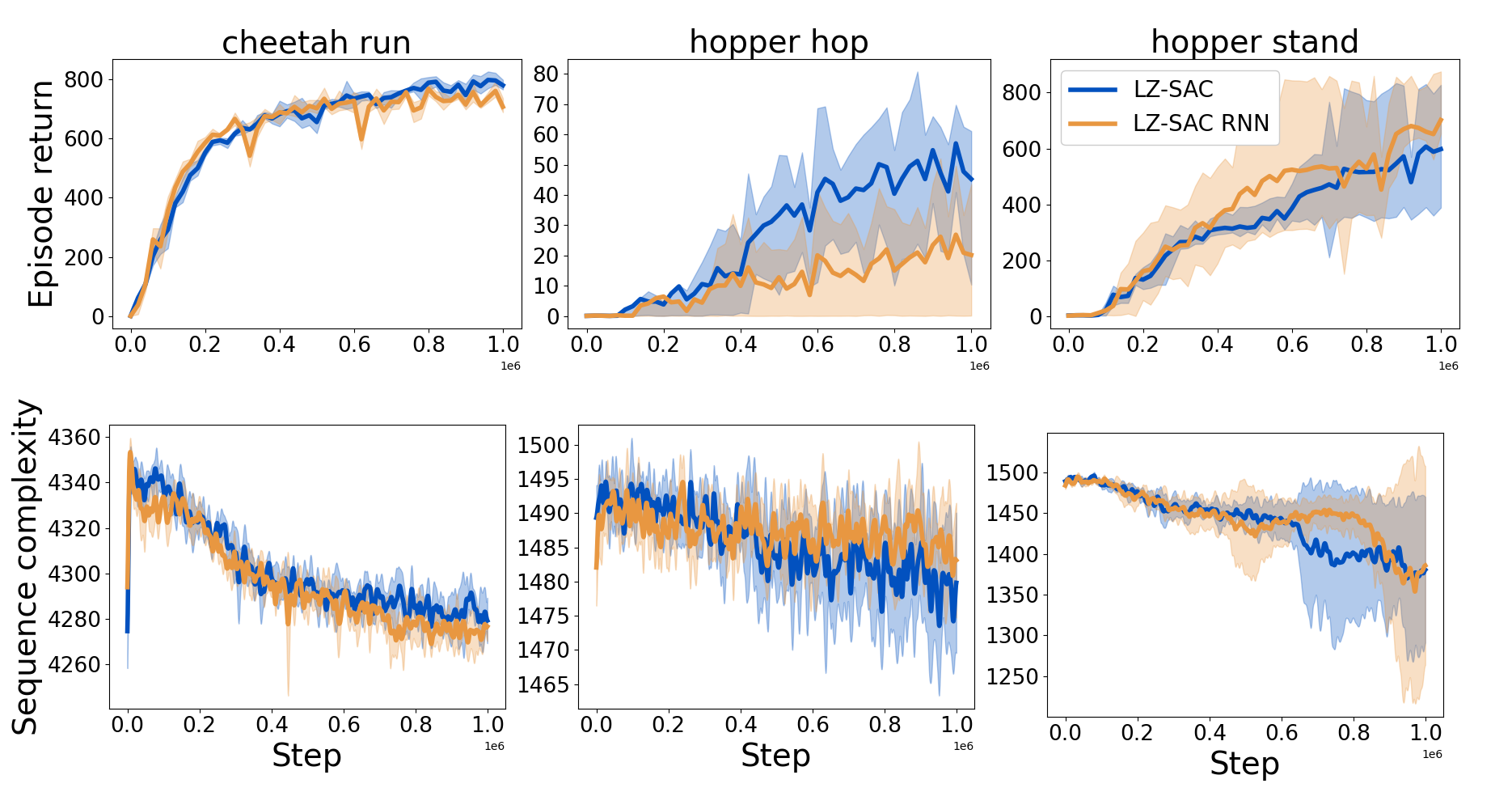}

\end{center}
\caption{Returns and compressibility of action sequences when the reward function is partially observable and fully observable.}\label{fig:appendix_augmented}

\end{figure}

The augmented reward function that induces the preference for simple action sequences depends on the actions the agent selected in the past. This makes the reward function partially observable for a purely state-conditioned policy. Our agents learn to maximize this reward function despite this partial observability. We tested whether augmenting the state to contain information about actions selected in the past produced substantial differences in the learned policies. We equipped the LZ-SAC agents with a recurrent neural network (a Gated Recurrent Unit \cite{chung2014empirical}) whose inputs were sequences of actions. We trained this network along with a single readout layer to produce embeddings $\mathbf{e}_t$ of action sequences with which we defined the augmented state $\action\sim\policy(\cdot|\tilde{\mathbf{s}}_t)$ where $\tilde{\mathbf{s}}_t = Concatenate(\state, \mathbf{e}_t)$. In three tasks we observed only minor differences in the policies learned (Fig \ref{fig:appendix_augmented}).   

\newpage

\end{document}